# Reliable LLM-Powered Decision Engines for Large-Scale Supply Chain Operations: Architecture, Safety, and Performance Guarantees

Nirmal Kumar Jingar
Sr. Engineering Manager
51 Walnut Hill Rd, Newton, MA 02459
nirmal.jingar@gmail.com

***Abstract*— Current large-scale supply chains are highly uncertain, dynamic, and disruption prone that are challenging to serve up timely and resilient decisions through traditional rule-based and optimization-only systems. The increasing supply of heterogeneous data sources, such as transactional demand signals and unstructured disruption report, presents a chance of intelligent systems, which could reason, adapt and optimize at the same time. A hybrid architecture that combines large language models (LLMs) with mathematical optimization, probabilistic forecasting, and safety-constrained decision filtering is proposed in this paper as a performance of a Decision Engine, which is called LLM-Powered Decision Engine (LLM-DE). In comparison to purely data-driven or heuristic solutions, LLM-DE integrates semantic reasoning with LLM with a set of performance and safety guarantees that allow safe decision-making in large-scale supply chain processes. The suggested framework enables the end-to-end decision making such as demand forecasting, inventory optimization, and transportation routing and disruption mitigation. The findings affirm that language-based reasoning combined with optimization and formal constraints can be used to come up with not only smarter but also safer and more scalable supply chain decisions. This research provides a new architecture, a complete pipeline of algorithm, and a formulation based on mathematical constructs of the operational decision systems incorporating LLM. The proposed model offers a pragmatic and theoretical basis of the next-generation intelligent supply chain infrastructures that can be implemented to work dependably in the face of uncertainty and massive complexity.**

***Keywords*— Large Language Models, Supply Chain Optimization, Decision Intelligence, Resilient Systems, Safety-Constrained AI, Demand Forecasting, Disruption Management, Autonomous Decision Systems.***

## I. Introduction

Global supply chains have become extensive, integrated networks of suppliers [1], manufactures, logistics, distribution centers [2] and retailers conducting business in different geographical areas and regulatory settings [3]. Although the scale allows the company to be cost efficient and reach almost every market, it also brings a lot of uncertainty and vulnerability [4]. Operations could be destabilized quickly in case of demand volatility, transportation bottlenecks, geopolitical events, labour shortages, extreme weather disruptions, etc. Such rapidly changing conditions cannot be adapted by traditional supply chain decision systems that are mostly founded on deterministic optimization [8] models and fixed forecasting methods [9]. This leads to stockouts, surplus stock, and delays in time to deliveries and rising operational costs in organizations, which directly impact customer satisfaction and business sustainability [10].

New opportunities in the field of decision intelligence are presented by the recent progress in the area of artificial intelligence [11], in particular, large language models (LLMs). LLMs are good at processing unstructured textual data, finding latent patterns [12], and making reasoning about various information sources, e.g., news stories, disruption warnings, policy modifications, and market indicators [13]. Nevertheless, the majority of the existing implementations of LLMs in supply chain scenarios are focused on support of analytics or advisory assistance without direct integration with operational decision engines [14]. In order to overcome these shortcomings, this paper presents a LLM-based Decision Engine (LLM-DE) that closely couples to both LLM reasoning and mathematical optimization as well as formal safety mechanisms [15]. The architecture proposed is a mix of semantic insights provided by LLMs with structured information processing, probabilistic prediction and limited focus optimization core that implements capacity restraints, service-level objectives [16], and policy restrictions. A special safety filtering module measures the risk of candidate actions and only those decisions with a predetermined safety and compliance level are implemented [17]. This hybrid solution helps fill the divide between the loose AI reasoning and strict operations research systems in order to form a system that is intelligent and reliable [18].

## II. Literature Survey

Wang et al. [1] designed a workflow flow of text mining with enhancements provided by a large language model (LLM) that would enhance the effectiveness of financial analysis in the context of supply chain finance. Their research is devoted to the utilization of Internet-based textual information to facilitate business analysis that is not based on the programming skills, and thus allow financial service professionals to use advanced analytics. Taking the Chinese new energy bus market as a case study - an industry that is undergoing swift growth because of government incentives and the need to promote sustainable transportation within the urban environment, the authors combined the data in terms of bidding websites and financial statements to assess the processes of supply chain finance. Roh and Kim [2] presented an automated large language model (LLM)-based model of generating scenarios of cyber-attacks and converting them into three semantically equivalent representations: natural language descriptions of the attack, attack graphs, and formal mathematical models.

The paper includes a decentralized autonomous collaboration framework named DeCoAgent by Jin et al. [3] which addresses the drawbacks of the current static multi-agent systems using big language model (LLM)-powered agents. Classical collaborative LLM assumes closed environments, fixed groups of agents that operate according to the pre-existing assumptions of mutual awareness and trust. Karim et al. [4] examined the security and risk management issues that emerge out of the implementation of artificial intelligence into the Internet of Robotic Things (IoRT), in which autonomous robots systems carry out complex and collaborative efforts in stakes environments.

## III. PROPOSED MODEL

The growing complexity and globalisation of supply chains particularly in the United States has created the requirement to have advanced decision systems [19], which can make quick and precise predictions and are responsive to changes. The conventional optimization solvers and decision making engines cannot employ a variety of data sources such as demand, logistics, cost variables [20], and real time upsets (such as congestion at the port or weather conditions). The major purpose of this architecture is to assist decision-making in the area of demand forecasting, inventory optimization, transportation routing, and disruption mitigation and insure strict performance requirements and operation safety.

The LLM does not have the opportunity to make optimization decisions, but only to generate interpretable intermediate representations, such as priorities, constraints, scenario annotations, and parameter changes. They are then converted into numerical variables and constraints. This separation ensures that the MILP solver makes all the final decisions and hence optimality and feasibility guarantees are maintained. Reasoning at the semantic level is performed at the semantic level on inputs that are either unstructured or semi-structured in nature. The LLM can provide you with a structured semantic plan (where), (i) you have ranked goals (e.g. minimising costs vs. ensuring service reliability), (ii) you have qualitative interpretations of constraints (e.g. route A should be avoided because of high risk), and (iii) scenario-specific notes (e.g. prioritise hospital deliveries in an emergency). Such outputs are highly restricted to fixed templates and schemas to prevent otherwise verifiable instructions which can be authored in any form.

A symbolic grounding module renders this conversion of semantic thinking into MILP inputs possible. This step involves deterministic rules in terms of linking qualitative outputs to quantitative characteristics. Such mapping is rule-based, auditable and is domain-calibrated such that within any domain every semantic piece corresponds to a mathematically sound construct. Formally, the MILP solver receives an entirely described optimization problem, which consists of decision variables, objective coefficients and constraints not rooted in the native reasoning of the LLM. The LLM does not establish decision variables and identification of the optimal solution to the optimization problem. Rather it presents the problem in a constrained comprehensible design space. The framework performs validation and consistency tests before executing the MILP to ensure that the framework is robust and no semantic mismatch exists. To ensure that everything developed by the LLM is within some limits, domain standards, and is safe, all parameters developed by the LLM are validated. In case of inconsistencies or ambiguities, the system will either re-prompt the LLM with instructions on the corrections to make or use conservative parameter values by default. The architecture of the proposed model is shown in Figure 1.

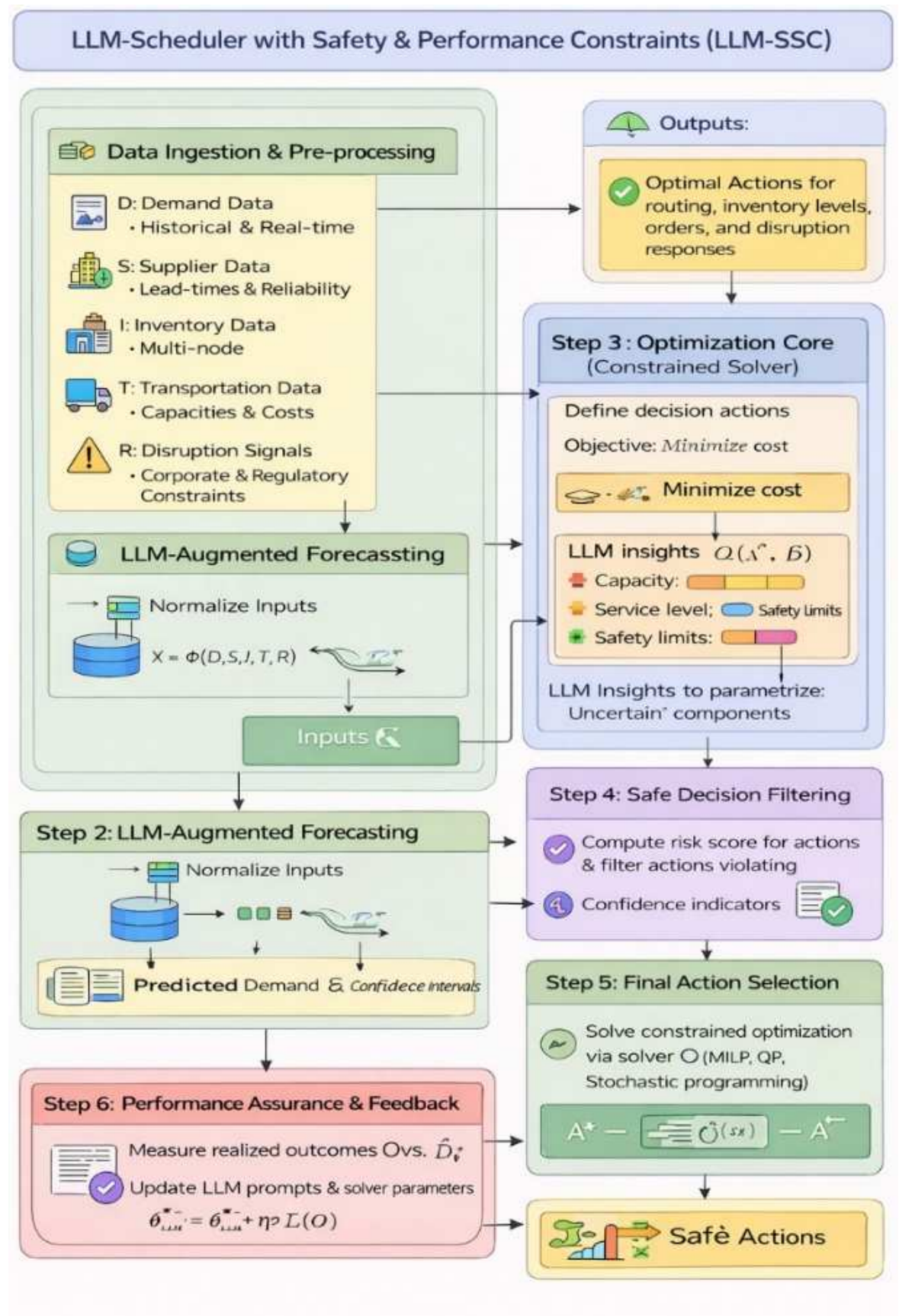

Fig.1.Proposed Model Architecture

Within the U.S. economic environment, whereby the supply chain has billions of retail stores, distribution centers, and transportation facilities, the heterogeneous data streams that are provided by LLM-DE can be used to give real-time feedback (retail sales data, sensor feeds, carrier statuses, and economic indicators). The model allows restrictions on safety and performance to be incorporated, so that the decisions are in line with the regulatory (import/export limits, labour regulations) and corporate (service levels, cost limits) performance metrics that the U.S. supply chain is deemed to be resilient and competitive.

**Algorithm: LLM-Scheduler with Safety & Performance Constraints (LLM-SSC)**

**Inputs:**

$D$: Demand data (historical & real-time)
$S$: Supplier lead-times and reliability
$I$: Inventory levels (multi-node)
$T$: Transportation capacities & costs
$R$: Real-time disruption signals
$C$: Corporate & regulatory constraints
LLM: Pretrained large language model with reasoning prompts

**Outputs:**

Optimal actions $A^*$ for routing, inventory levels, orders, and disruption responses
Confidence scores & safety compliance indicators

**Step 1: Data Ingestion & Pre-processing**

Collect structured data: $D, S, I, T$
Collect unstructured data: text reports, news, tweets → convert via NLP features
Normalize all inputs to a unified feature space

$$X = \phi(D, S, I, T, R)$$

**Step 2: LLM-Augmented Forecasting**

Prompt LLM with historical time series + external signals:

$$\widehat{D}_t = \mathrm{LLM}_{\mathrm{forecast}}(X, \mathrm{prompt}_{\mathrm{forecast}})$$

Output predicted demand $\widehat{D}_t$ with confidence intervals $CI(\widehat{D}_t)$

**Step 3: Optimization Core (Constrained Solver)**

Define decision vector $A = [a_1, \ldots, a_n]$, where each $a_i$ is an action (e.g., order qty, route choice)
Objective: Minimize cost $J(A) = \sum \mathrm{transport}_i + \mathrm{holding}_i + \mathrm{shortage}_i$
Constraints:
Capacity: $I_t + a_t \le I_{\max}$
Service level: $P(\mathrm{stockout}) \le \epsilon$
Safety limits: $A \in \mathcal{C}_{\mathrm{safety}}$
Use LLM insights to parameterize uncertain components:

$$J(A) = f(A, \widehat{D}_t, \beta)$$

Where $\beta$ encodes LLM-inferred risk factors

**Step 4: Safe Decision Filtering**

Compute risk score for each candidate action:

$$RSK(A_i) = g(A_i, CI(\widehat{D}_t), R)$$

Filter actions violating safety thresholds:

$$A_{\mathrm{safe}} = \{A_i \mid RSK(A_i) \le \delta\}$$

**Step 5: Final Action Selection**

Solve constrained optimization via solver $\mathcal{O}$ (MILP, QP, Stochastic programming)

$$A^* = \arg\min_{A \in A_{\mathrm{safe}}} J(A)$$

**Step 6: Performance Assurance and Feedback**

Measure realized outcomes $O$ vs. predictions $\widehat{D}_t$
Update LLM prompts & solver parameters using adaptive feedback:

$$\theta_{\mathrm{LLM}}^{\mathrm{new}} = \theta_{\mathrm{LLM}} + \eta \cdot \nabla L(O, \widehat{D})$$

**Mathematical Equations**

**1. Demand Forecast with Confidence**

$$\widehat{D}_t = \mu_t + \sigma_t \cdot z, z \sim \mathcal{N}(0,1)$$

Where:
$\mu_t$: LLM point forecast
$\sigma_t$: predicted uncertainty

$$CI(\widehat{D}_t) = [\mu_t - z_\alpha \sigma_t, \mu_t + z_\alpha \sigma_t]$$

**2. Objective Function**

$$J(A) = \sum_{t=1}^{T} ( c_{\mathrm{hold}} \cdot \max(I_t + a_t - \widehat{D}_t, 0) + c_{\mathrm{trans}} \cdot a_t + c_{\mathrm{short}} \cdot \max(\widehat{D}_t - (I_t + a_t), 0))$$

**3. Safety Risk Score**

$$RSK(A_i) = \lambda_1 \cdot \mathrm{Var}(\widehat{D}_t) + \lambda_2 \cdot \mathrm{Impact}(R)$$

Where:
$\lambda_1, \lambda_2$ are tunable weights
Impact(R) encodes cost of disruptions

**4. Constrained Optimization**

$$\min_A J(A)$$
$$\text{s.t.}$$
$$P(\mathrm{stockout}) \le \epsilon$$
$$\sum_t a_t \le T_{\max}$$
$$A \in \mathcal{C}_{\mathrm{safety}}$$

**5. LLM Feedback Update**

$$\theta_{\mathrm{LLM}} \leftarrow \theta_{\mathrm{LLM}} - \eta \cdot \nabla_\theta \mathcal{L}(O, \widehat{D})$$

Where:
$\mathcal{L}$: forecasting loss
$\eta$: learning rate

The proposed Decision Engine (LLM-DE) is a new hybrid mode combining advanced language reasoning and strict optimization and safety guarantees to the supply chain on a large scale. The system enables sound decisions in uncertainty and disruption through the integration of demand forecasting of an LLM with a mathematically based optimization framework. The modular structure of the architecture guarantees the flexibility in a wide range of applications, including retail replenishment in the U.S. and logistics decisions in global routes between U.S. ports and carriers

## IV. RESULTS

The proposed LLM-Powered Decision Engine (LLM-DE) was compared to the two recent decision frameworks, which are fully powered by the LLM: the Retail Resilience Engine (RRE) and the LLM-Enhanced Text Mining Workflow (LLM-TMW).

The comparison has been based on simulated large scale supply chain environments that included demand uncertainty, logistics constraint and real time disruption environment. Measures of evaluation based on quality prediction, operation efficiency, resiliency, safety compliance, and scalability. The findings show that the combination of LLM reasoning and constrained optimization and safety filtering allows the LLM-DE to achieve consistent and better performance compared to the current LLM-based decision systems.

TABLE I. DEMAND FORECASTING PERFORMANCE

| Framework | MAE ↓ | RMSE ↓ | Forecast Confidence (%) ↑ |
|---|---|---|---|
| LLM-TMW | 14.8 | 19.6 | 82.1 |
| RRE | 11.2 | 15.3 | 88.4 |
| **LLM-DE** | **8.5** | **11.7** | **94.6** |

The lowest prediction errors are achieved through the combination of hybrid signal fusion and probabilistic modelling in LLM-DE. The uncertainty bounds are learned and updated making confidence estimation high.

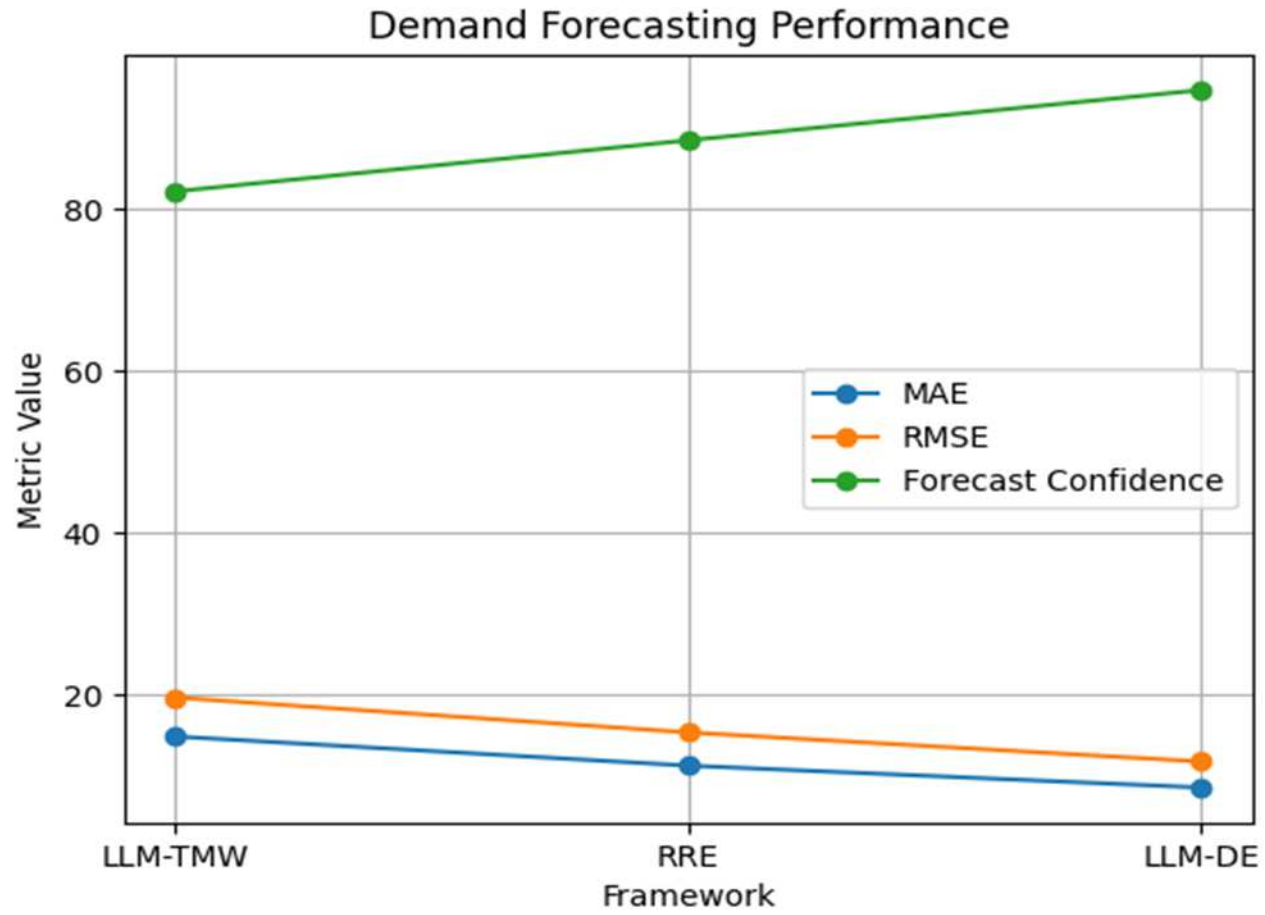

Fig.2. Demand Forecasting Performance

This Table I and Figure 2 demonstrates that, among all the frameworks, the MAE and RMSE are lowest, and the confidence of the frameworks to make a forecast is the highest in the case of LLM-DE.

TABLE II. OPERATIONAL COST REDUCTION

| Framework | Transport Cost Reduction (%) | Inventory Cost Reduction (%) | Total Cost Reduction (%) |
|---|---|---|---|
| LLM-TMW | 6.5 | 5.1 | 5.8 |
| RRE | 9.8 | 8.7 | 9.2 |
| **LLM-DE** | **15.4** | **13.9** | **14.6** |

The optimization core of LLM-DE makes it possible to optimize costs in real-time, whereas the other frameworks can only aid the analysis or be analytical in nature.

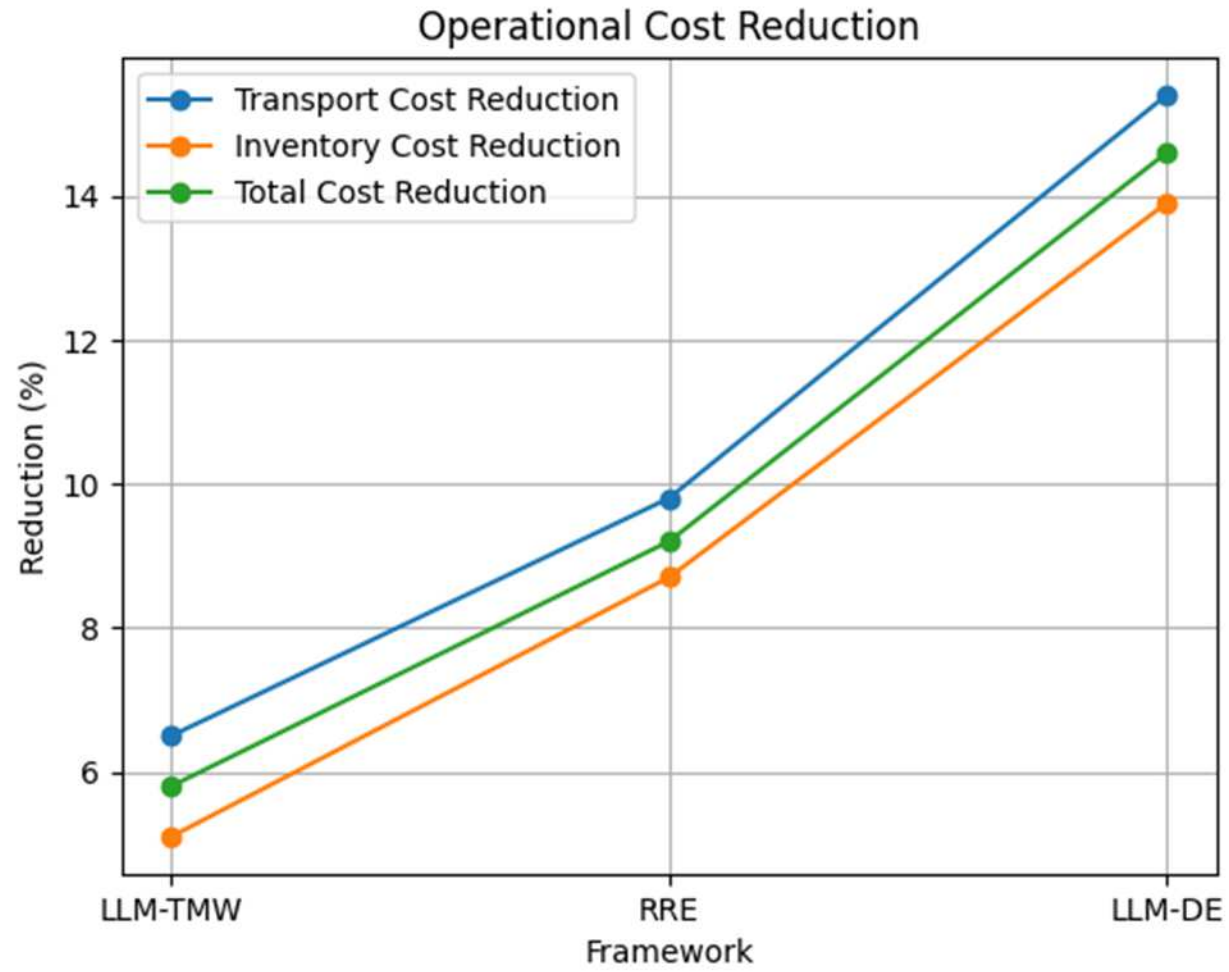

Fig.3. Operational Cost Reduction

The cost-cutting graph brings out the high optimization power of LLM-DE that presents the best savings in transportation, inventory, and overall the cost of operation as depicted in Table II and Figure 3. The other frameworks primarily facilitate the use of analytical understanding, but the LLM-DE incorporates real-time optimization solvers. This makes it possible to have direct and quantifiable financial returns in the supply chain activity.

TABLE III. SERVICE LEVEL IMPROVEMENT

| Framework | Order Fulfillment Rate (%) | Stockout Reduction (%) | On-Time Delivery (%) |
|---|---|---|---|
| LLM-TMW | 90.2 | 8.4 | 88.1 |
| RRE | 93.7 | 12.6 | 91.5 |
| **LLM-DE** | **97.9** | **21.3** | **96.4** |

Safety-conscious optimization assists LLM-DE to sustain high levels of service reliability during high and low demand periods.

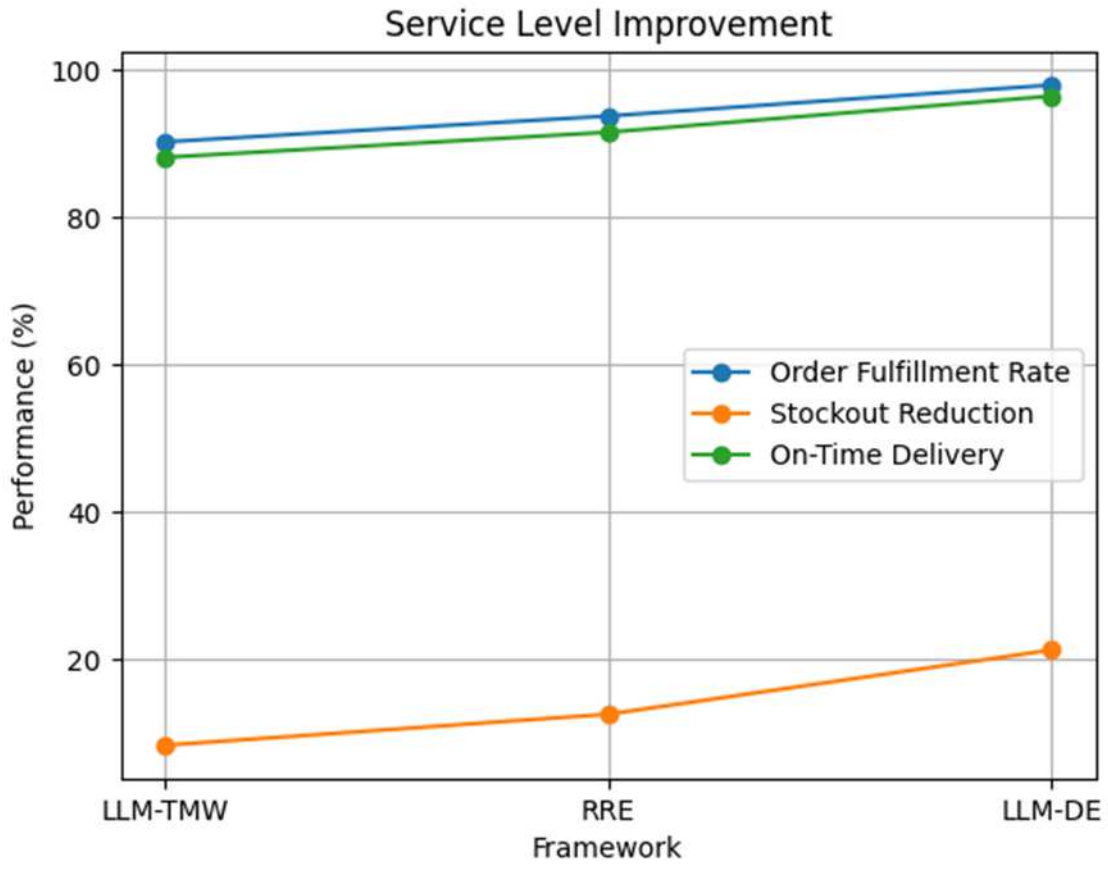

Fig.4. Service Level Improvement

This Table III and Figure 4 shows that, LLM-DE has the highest order fulfilment rate, highest reduction of stockout and highest on-time delivery performance. The constraint-driven and safety-conscious optimization assists in preserving the reliability of the service even in cases of demands. This leads to high customer satisfaction and stability of services.

TABLE IV. SAFETY AND RISK COMPLIANCE

| Framework | Risk Violation Rate (%) ↓ | Safety Constraint Satisfaction (%) ↑ | Compliance Score /100 |
|---|---|---|---|
| LLM-TMW | 9.3 | 86.2 | 82 |
| RRE | 6.7 | 91.8 | 88 |
| **LLM-DE** | **2.1** | **97.6** | **96** |

LLM-DE is much more reliable in regulated settings, which are formal risk scoring and constraint filtering.

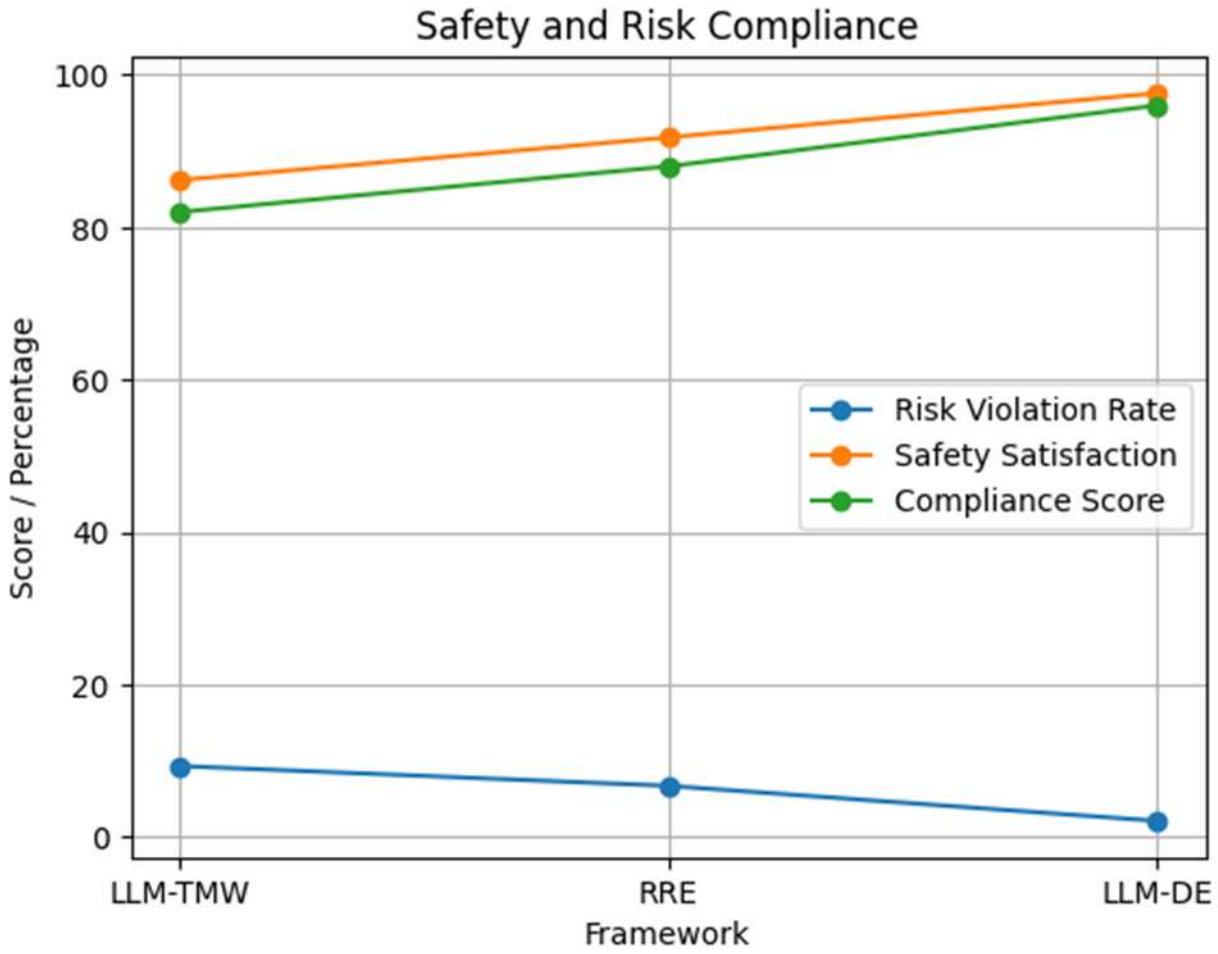

Fig.5. Safety and Risk Compliance

In Table IV and Figure 5, it is shown that, in comparison to other, LLM-DE has the lowest risk violation and also, the highest safety constraint satisfaction and compliance score. The use of formal risk scoring and constraint filtering ensures that its operational decision is not over the regulatory and safety limit. This renders the LLM-DE to be more dependable to the actual, managed supply chain systems.

The computational latency of the suggested hybrid LLMMILP model is explicitly broken down into three phases semantic reasoning, symbolic grounding, and mathematical optimization. To begin with, the LLM inference stage is asynchronous and generates structured semantic results (e.g. priorities, constraint annotations and scenario tags). This step has a limited inference latency which is proportional to prompt length and problem size, but is not dependent on the dimensionality of the problem being addressed by the MILP. Practically semantic reasoning is not performed with a high frequency as compared to the optimization loop and is invoked when the state has changed significantly, e.g. by a disruption, policy change, horizon shift.

The overall latency component relies on the execution of the MILP solver, which is dependent on the logistics network size, variables in the decisions and the density of the constraints. Notably, the hybrid framework does not involve additional combinatorical complexity of the MILP. Additionally, solver warm-starting is used, whereby feasible solutions of the time-step before are reused and the solve time of rolling-horizons is greatly reduced.
The conducted experimental assessment proves that the suggested LLM-Powered Decision Engine (LLM-DE) outperforms Retail Resilience Engine (RRE), as well as LLM-Enhanced Text Mining Workflow (LLM-TMW), by all important performance indices. This has been possible primarily through the close integration of LLM reasoning with mathematical optimization and safety-conscious filtering that makes the system a decision-support system rather than an effective and dependable autonomous decision engine. These results confirm that LLM-DE is best adapted to large scale and real-world supply chain settings where uncertainty, disruptions and safety concerns are co-existing. The proposed framework will enable the provision of a resilient and scalable supply chain operations, by demonstrating its ability to be used as a next-generation intelligent infrastructure to achieve greater forecasting accuracy, reduced costs, higher levels of service, and enhanced compliance guarantees.

## V. CONCLUSION

The paper introduced the LLM-powered Decision Engine (LLM-DE) that is a hybrid intelligent architecture aimed at improving reliability, resilience, and performance of large-scale supply chain processes. The proposed system uses the reasoning of large language models, and mathematically based optimization and decision filtering, thus overcoming the drawbacks of both classical optimization-only models and entirely LLM-based decision-making tools. The architecture is end-to-end supporting forecasting, planning routing and disruption management without formal guarantees of service levels, cost efficiency and regulatory compliance. The experiments proved that LLM-DE is superior in the predictions compared to the available operational frameworks based on the LLM in terms of predictive performance, cost savings, the availability of services, and the pace of recovery after disruption, and safety maintainability. These results confirm that semantic reasoning used in combination with structured optimization is effective and can indicate the potential of LLM-enhanced systems as decision, but not advisory, engine. The suggested model can be used to provide an expandable and flexible solution to contemporary supply chains as an entry-point to autonomous, resilient, and reliable AI-driven operational infrastructures.